\documentclass[11pt]{article}

\usepackage[preprint]{acl}

\usepackage{times}
\usepackage{latexsym}

\usepackage[T1]{fontenc}

\usepackage[utf8]{inputenc}

\usepackage{microtype}

\usepackage{inconsolata}

\usepackage{graphicx}

\usepackage{amsmath,hyperref}
\usepackage{multirow} 
\usepackage{booktabs}

\usepackage{minted}
\usepackage{enumitem}
\usepackage{amssymb}

\usepackage{tcolorbox}
\tcbuselibrary{breakable,skins}
\usepackage{fvextra}
\usepackage{wrapfig}
\newtcolorbox{promptbox}{
  enhanced jigsaw,
  breakable,                   
  colback=black!2,            
  colframe=black!30,
  boxrule=0.5pt, arc=2mm,
  left=1mm,right=1mm,top=1mm,bottom=1mm
}

\title{LLM-Powered Text-Attributed Graph Anomaly Detection via Retrieval-Augmented Reasoning}

\author{
\textbf{Haoyan Xu\textsuperscript{1}$^{*}$},
\textbf{Ruizhi Qian\textsuperscript{1}$^{*}$},
\textbf{Zhengtao Yao\textsuperscript{1}},
\textbf{Ziyi Liu\textsuperscript{1}},
\textbf{Li Li\textsuperscript{1}},
\textbf{Yuqi Li\textsuperscript{3}},
\textbf{Yanshu Li\textsuperscript{4}},\\
\textbf{Wenqing Zheng\textsuperscript{2}},
\textbf{Daniele Rosa\textsuperscript{2}},
\textbf{Daniel Barcklow\textsuperscript{2}},
\textbf{Senthil Kumar\textsuperscript{2}},
\textbf{Jieyu Zhao\textsuperscript{1}},
\textbf{Yue Zhao\textsuperscript{1}}
\\
\textsuperscript{1}University of Southern California,
\textsuperscript{2}Capital One,
\textsuperscript{3}City College of New York,
\textsuperscript{4}Brown University
\\
\texttt{haoyanxu@usc.edu, ruizhiqi@usc.edu, zyao9248@usc.edu, zliu2803@usc.edu,}\\
\texttt{li.li02@usc.edu, yuqili010602@gmail.com, yanshu\_li1@brown.edu}
\\
\texttt{\{wenqing.zheng, daniele.rosa, daniel.barcklow, senthil.kumar\}@capitalone.com} \\
\texttt{jieyuz@usc.edu, yzhao010@usc.edu}
\\
}

\begin{document}
\maketitle

\def\thefootnote{*}\footnotetext{Equal contribution.}
\def\thefootnote{\arabic{footnote}}

\begin{abstract}
Anomaly detection on attributed graphs plays an essential role in applications such as fraud detection, intrusion monitoring, and misinformation analysis. However, text-attributed graphs (TAGs), in which node information is expressed in natural language, remain underexplored, largely due to the absence of standardized benchmark datasets. In this work, we introduce TAG-AD, a comprehensive benchmark for anomaly node detection on TAGs. TAG-AD leverages large language models (LLMs) to generate realistic anomalous node texts directly in the raw text space, producing anomalies that are semantically coherent yet contextually inconsistent and thus more reflective of real-world irregularities. In addition, TAG-AD incorporates multiple other anomaly types, enabling thorough and reproducible evaluation of graph anomaly detection (GAD) methods. With these datasets, we further benchmark existing unsupervised GNN-based GAD methods as well as zero-shot LLMs for GAD.


As part of our zero-shot detection setup, we propose a retrieval-augmented generation (RAG)-assisted, LLM-based zero-shot anomaly detection framework.
The framework mitigates reliance on brittle, hand-crafted prompts by constructing a global anomaly knowledge base and distilling it into reusable analysis frameworks. Our experimental results reveal a clear division of strengths: LLMs are particularly effective at detecting contextual anomalies, whereas GNN-based methods remain superior for structural anomaly detection. Moreover, RAG-assisted prompting achieves performance comparable to human-designed prompts while eliminating manual prompt engineering, underscoring the practical value of our RAG-assisted zero-shot LLM anomaly detection framework.

\end{abstract}

\section{Introduction}
\label{sec:intro}
Graph anomaly detection (GAD) has become a critical task in graph mining \cite{ding2019deep, fan2020anomalydae, xu2024lego, xu2025few, xu2025glip, xu2025graph}, with applications spanning fraud detection, cybersecurity, and scientific discovery. Existing research has primarily focused on attributed graphs \cite{liu2021anomaly, xu2021graph}, where node features are represented as fixed-dimensional vectors. However, in many real-world scenarios, nodes are naturally described by textual attributes (e.g., paper abstracts in citation networks \cite{giles1998citeseer}, user bios in social graphs \cite{xiao2020timme}, or product descriptions in e-commerce). Despite their prevalence, text-attributed graphs (TAGs) remain largely underexplored for anomaly detection, primarily due to the absence of publicly available benchmark datasets, which hinders systematic comparison and evaluation of emerging methods.

In this work, we address this gap by constructing the first comprehensive text-attributed graph anomaly detection (TAG-AD) benchmark dataset. Our key idea is to leverage large language models (LLMs) to directly generate realistic anomalous node texts in the raw text space. Compared with previous methods that inject anomalies by perturbing node features in the embedding or feature space, generating anomalies in the raw text space is more flexible, offers better interpretability, and preserves the fluency and semantic coherence of node attributes. Beyond LLM-generated anomalies, our framework also incorporates multiple additional types of contextual and structural anomalies.
We instantiate this pipeline on several widely used real-world graph datasets (e.g., Cora, CiteSeer, etc.), producing a curated benchmark that reflects both structural and contextual anomaly patterns.

\begin{figure}[t]
    \centering
    \includegraphics[width=1.0\linewidth]{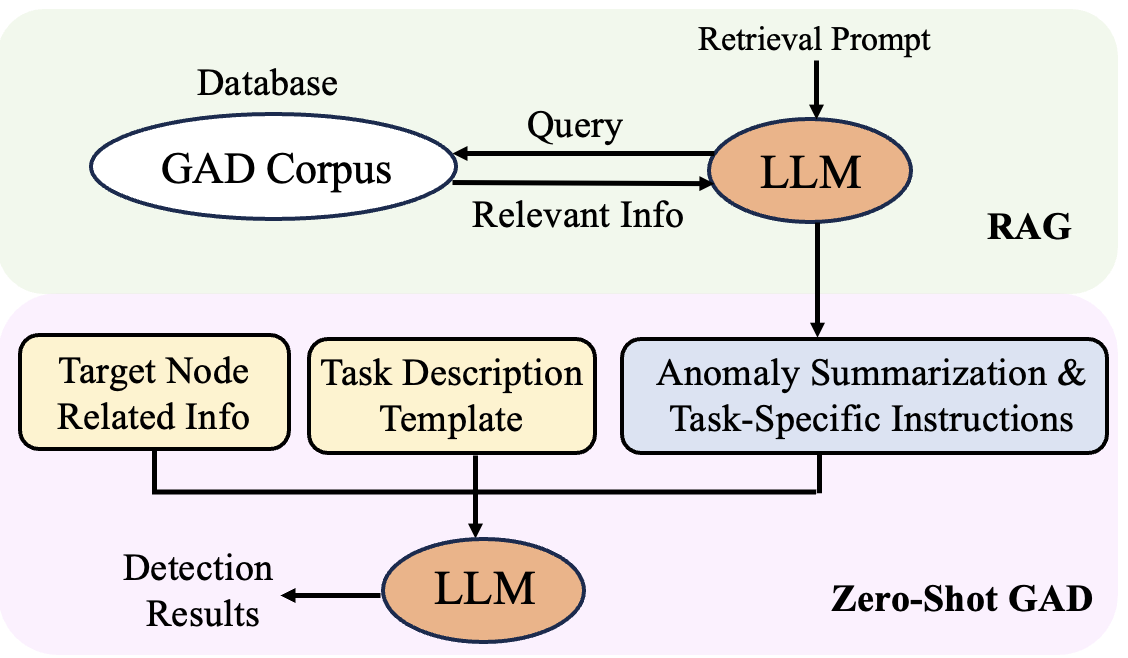}
    \caption{Our method introduces a RAG framework for zero-shot GAD, in which a globally retrieved anomaly knowledge base is distilled by an LLM into unified detection guidelines. During inference, each prompt integrates this analysis framework, the task description, and node-specific graph context to enable consistent and interpretable anomaly reasoning.}
    \label{fig:TAG-AD}
    \vspace{-0.15in}
\end{figure}

To evaluate the utility of this benchmark, we conduct a comprehensive assessment of state-of-the-art unsupervised GNN-based GAD methods on the proposed TAG-AD dataset. Furthermore, we explore the zero-shot anomaly detection capabilities of modern LLMs (e.g., Qwen3-14B \cite{yang2025qwen3}, GPT-4o-mini \cite{hurst2024gpt}, Deepseek-V3 \cite{liu2024deepseek}, Gemma-3-27b-it \cite{gemmateam2025gemma3technicalreport}) by prompting them directly with node texts and local graph structure information, offering a new perspective on how LLMs can serve as anomaly detectors without task-specific training. In total, our benchmark includes four datasets, each containing five types of anomalies, and we evaluate six unsupervised GNN-based methods and four LLMs across all settings.

A practical limitation we observe is that zero-shot GAD with LLMs often relies on hand-crafted prompts that must encode detailed knowledge about the specific types of anomalies present in a graph. Moreover, LLMs require explicit instructions to accurately detect diverse types of anomalies across graphs from different domains. Such prompt engineering is brittle, time-consuming, and poorly scalable across datasets and anomaly categories in TAGs. To address this limitation, we introduce a retrieval-augmented generation (RAG) \cite{lewis2020retrieval, gao2023retrieval} scheme tailored for zero-shot GAD. In our design, the retrieval process is performed once at the global level, independent of any specific node or dataset, and provides a shared knowledge base summarizing canonical definitions and taxonomies of anomaly types (e.g., contextual, structural, textual-semantic). The retrieved content is distilled by the LLM into concise detection guidelines, which form a unified \textit{analysis framework}.

As shown in Fig. \ref{fig:TAG-AD}, during inference, each prompt combines three components: (i) the RAG-generated analysis framework that provides anomaly definitions and reasoning criteria; (ii) a task description specifying the detection task and scoring rubric (e.g., assigning a 0-10 anomaly score); and (iii) node-specific information, including the target node’s text and its neighborhood or structural context. This structured prompt ensures that each decision is grounded in retrieved anomaly knowledge, guided by consistent evaluation rules, and informed by the target node’s local graph context. The RAG-assisted prompting design decreases the reliance on manual prompt engineering and substantially improves robustness and consistency across datasets and anomaly categories. In summary, our contributions are threefold:
\begin{itemize}[leftmargin=*, itemindent=0pt]
\item We construct the first comprehensive TAG anomaly detection dataset, using LLMs to generate realistic node-level anomalies in the raw text space and incorporating additional contextual and structural anomaly types.

\item We propose a RAG-assisted prompting framework for zero-shot GAD that retrieves and summarizes anomaly knowledge and detection instructions from the literature, replacing manually designed prompts with a structured combination of the Analysis Framework, the Task Description and Scoring Rubric, and the Node-specific Information.

\item We systematically benchmark a variety of unsupervised GAD methods under the TAG setting and investigate the zero-shot anomaly detection ability of LLMs, highlighting their strengths and limitations relative to graph-based methods. We release the testbed, corresponding code, and the proposed pipelines at: \url{https://github.com/Flanders1914/TAG_AD}. 
\end{itemize}

We hope this work provides a foundation for advancing GAD research on TAGs and opens new directions for integrating graph learning with foundation language models.

\section{Related Work}
\label{sec:related}
\subsection{Text-Attributed Graph Datasets}
TAG datasets have become a cornerstone for advancing research at the intersection of graph learning and LLMs. The recent surge in interest has led to the development of a diverse array of benchmarks. \cite{yan2023comprehensive} introduces CS‑TAG, a diverse and large-scale suite of benchmark datasets for TAGs, and establishes standardized evaluation protocols. Recognizing the importance of temporal dynamics, \cite{NEURIPS2024_a65d054a} introduced DTGB, a dynamic TAG dataset. \cite{10.5555/3737916.3739865} further proposed TEG-DB, which incorporates both node and edge textual attributes. Despite these advances, there is no comprehensive TAG dataset for anomaly detection due to a lack of anomalies.

\subsection{Node-level Anomaly Detection}
Anomaly detection on graphs has been studied extensively in the context of attributed graphs, where node features are represented as fixed-dimensional vectors. Representative approaches include reconstruction-based methods such as DOMINANT \cite{ding2019deep}, Autoencoder-based detectors like AnomalyDAE \cite{fan2020anomalydae}, contrastive methods such as CoLA \cite{liu2021anomaly}. These methods demonstrate strong performance on standard attributed graph benchmarks, but they are not directly applicable to TAGs where node attributes are texts. Existing methods typically rely on pre-computed embeddings, which may fail to capture fine-grained textual inconsistencies characteristic of contextual anomalies in TAGs. Recently, a few studies have explored AD on TAGs \cite{xu2025court, xu2025textattributedgraphanomalydetection}, but the datasets they used are quite limited. They rely on perturbing original textual attributes, which often \textbf{produces unnatural or incoherent semantics}, and their evaluation methods are restricted in scope.

\subsection{LLMs for Graphs}
A fast-growing body of work studies how LLMs can assist or replace graph models across node classification, link prediction, graph classification, and reasoning. For TAG-oriented GNN-based anomaly detection, we adopt the \emph{LLMs-as-Prefix} pipeline \cite{ren2024survey}, where an LLM first processes text-enriched graph data and then provides either node embeddings or pseudo-labels to enhance subsequent GNN training. Representative approaches that use LLM-generated embeddings to support GNNs include G-Prompt \cite{huang2023prompt}, TAPE \cite{he2023harnessing}, and LLMRec \cite{wei2024llmrec}.
For LLM-based zero-shot GAD, we follow a tuning-free, \emph{LLMs-only} pipeline that designs graph-understandable prompts to directly leverage pretrained LLMs for graph-related reasoning without further training. Notable works in this line include NLGraph \cite{wang2023can}, GPT4Graph \cite{guo2023gpt4graph}, Graph-LLM \cite{chen2024exploring}, GraphText \cite{zhao2023graphtext}, and \emph{Talk Like a Graph} \cite{fatemi2023talk}.

\section{TAG-AD}
\label{sec:method}

\subsection{Problem Formulation}

Let $G=(V,E,T,\mathbf{X})$ denote a text-attributed graph, where $V$ is the set of nodes, $E \subseteq V \times V$ is the set of edges, and $T=\{t_v \mid v \in V\}$ is the collection of node attributes with each $t_v$ represented as free-form text (e.g., titles, abstracts, or user-generated content). These text attributes can be transformed into embeddings $\mathbf{X} = x_v \mid v \in V\}$ using a pretrained model~\cite{reimers2019sentence}. The goal of anomaly node detection is to identify a subset of nodes $V_{ano} \subset V$ whose textual or structural characteristics deviate significantly from the majority.
Formally, we define an \emph{anomalous node} as one that violates expected consistency across either:
\begin{itemize}[leftmargin=*, itemindent=0pt]
  \setlength{\itemsep}{0.5pt}
  \setlength{\parskip}{0.5pt}
  \setlength{\parsep}{0.5pt}
    \item \textbf{Contextual consistency:} the semantic content of the target node $v$ diverges from that of its neighborhood $\mathcal{N}(v)$ or global domain patterns.
    \item \textbf{Structural consistency:} the connectivity pattern of node $v$ deviates from community-level norms, potentially forming irregular subgraphs.
\end{itemize}

In TAG-AD, contextual anomalies are generated directly in the text space to ensure realism and semantic diversity. Given a normal node $(v, t_v)$ and its neighborhood context, our objective is to produce a modified text $\tilde{t}_v$ such that $(v, \tilde{t}_v)$ is indistinguishable in style and fluency from authentic nodes but contextually anomalous with respect to $G$. The anomaly detection task is then to learn a scoring function $f: V \rightarrow \mathbb{R}$ that assigns higher scores to nodes in $V_{ano}$ than to normal nodes, without access to anomaly labels during training.

\subsection{Anomaly Generation Method}
\label{subsec:anomaly_gen}
We propose an LLM-based approach to generate realistic contextual anomalies directly in the text space of TAGs. Unlike prior perturbation-based approaches \cite{xu2025textattributedgraphanomalydetection}, which modify existing textual attributes through insertion or replacement, our method leverages the generative capabilities of LLMs to produce fluent and coherent yet contextually inappropriate text that is semantically inconsistent with its neighbors or original context. In addition, our \textbf{TAG-AD} framework incorporates other established techniques for injecting contextual and structural anomalies, ensuring a comprehensive and versatile benchmark for evaluation.

\subsubsection{LLM-generated Contextual Anomaly}
Our approach consists of two stages: Target Label Selection and Text Generation. Given a node for anomaly injection, the framework first determines an appropriate target label for the anomalous text, and then generates a corresponding text attribute conditioned on this label. This design ensures that the generated anomalies remain in-distribution while still deviating from the original semantics and the semantics of neighboring nodes.

\paragraph{Target Label Selection} 
We employ a label elimination strategy to identify a suitable anomalous target label. Starting from the full set of candidate labels, the node’s original label is first removed. Then, labels frequently appearing in the node’s $k$-hop neighborhood are iteratively eliminated, beginning with the most frequent labels among the closest neighbors. This process continues until a single label remains, which is selected as the target label. This strategy ensures that the chosen label maximally diverges from both the node’s original textual attributes and the predominant semantics in its local neighborhood.

\paragraph{Text Generation} 
Given the selected anomalous target label, we design dataset-specific prompts to guide the LLM in rewriting the original node text. The prompts are carefully crafted so that the generated text (1) aligns topically with the target label, (2) departs significantly from the original semantics, (3) maintains the writing style and structural characteristics of the dataset, and (4) preserves the approximate length distribution of the original text.  

\subsubsection{Global Contextual Anomaly}
Another type of anomaly we generate is the global anomaly, for which our framework adopts a text-attribute replacement strategy. Within a dataset centered on a specific theme, we replace the target node’s text with another text of the same format and high-level topic domain, but originating from a class not present in the original dataset. In this way, the anomalous node appears topically consistent with the graph, yet remains \textbf{out-of-distribution from the perspective of class semantics}.

\subsubsection{Text Perturbation Anomaly}
Our framework adopts the text perturbation strategy of \cite{xu2025textattributedgraphanomalydetection} to generate text perturbation anomalies. For each target node, a set of $K_{\text{candidiate}}$ candidate nodes is randomly sampled, and the semantic distances between the target node and each candidate are computed using cosine similarity on their text attribute vectors. The candidate node with the lowest similarity is then selected as the source of anomalous information. A sequence of $l$ tokens is then sampled from the source text and replaces an equal-length segment of the target text, introducing semantically inconsistent content.

\subsubsection{Structural Anomaly}
To generate structural anomalies, our framework leverages PyGOD \cite{JMLR:v25:23-0963}. Specifically, $m$ nodes are randomly sampled and fully connected to form a clique. This procedure is repeated $n$ times to construct $n$ such cliques, thereby injecting anomalous structural patterns into the graph.

\subsection{GNNs and LLMs for AD on TAGs}
For GNN-based unsupervised GAD methods, we first transform the raw text of all nodes into embeddings $\mathbf{X}$ using Sentence-BERT \cite{reimers2019sentence}. Then, we use the adjacency information with the transformed node embeddings to train GAD models. After training, we directly derive the anomaly scores from the well-trained GAD models.

For LLM-based zero-shot GAD methods, we input each target node into the LLM along with its textual description, neighborhood information, and task instructions that prompt the model to determine whether the node is anomalous. The model’s output is then converted into numerical anomaly scores, from which we compute the final GAD performance. More details are in Section \ref{subsubsec:imple_details}.

\subsection{RAG-Assisted Prompting for Zero-Shot GAD}
A key challenge in applying LLMs to zero-shot GAD is the reliance on hand-crafted prompts. These prompts must explicitly encode knowledge about how different types of anomalies should be defined for a given graph (e.g., “what is structural anomaly in a financial network”), which makes them costly to construct and difficult to scale across datasets and anomaly categories in TAGs.

To address this limitation, we propose a RAG scheme specifically tailored for zero-shot GAD on TAGs. The pipeline is illustrated in Fig. \ref{fig:TAG-AD} and consists of three stages: corpus construction, retrieval and summarization, and grounded prompting.

\subsubsection{Corpus Construction}
We construct a domain-specific corpus consisting of research papers, survey articles, and dataset documentation related to anomaly detection, with emphasis on attributed graph and text-attributed graph anomaly detection. This corpus serves as the source of background knowledge for anomaly definitions and detection principles.

\subsubsection{Retrieval and Summarization}
Rather than performing node-level or dataset-specific retrieval, our method conducts a single global retrieval step that is shared across all experiments. The LLM is prompted to read the corpus and summarize key information about anomaly categories, their definitions, and common indicators. The resulting summary is distilled into a compact set of structured instructions $S$, which include:
\begin{itemize}[leftmargin=*, itemindent=0pt]
  \setlength{\itemsep}{0pt}
  \setlength{\parskip}{0pt}
  \setlength{\parsep}{0pt}
    \item \texttt{Anomaly knowledge}: concise descriptions of anomaly types such as contextual, structural, and textual-semantic anomalies;
    \item \texttt{Detection guidelines}: high-level criteria for identifying inconsistencies in text or structure.
\end{itemize}
These instructions remain fixed across all nodes and datasets, providing consistent background knowledge for zero-shot evaluation.

\subsubsection{Grounded Prompting} 
Our final prompting design unifies the retrieved anomaly knowledge, the user-defined task description, and node-specific information into a single structured prompt for zero-shot detection. Each prompt fed to the LLM contains three components:

(1) \textbf{RAG-generated instructions ($S$):} these are globally shared definitions and detection guidelines obtained from the retrieval and summarization stage, forming an \texttt{Analysis Framework} that describes what constitutes an anomaly (e.g., textual inconsistency, semantic deviation, or structural irregularity). 
(2) \textbf{Task description and scoring rubric:} the LLM is instructed to analyze the given node using the analysis framework and assign an integer anomaly score between 0 and 10, where higher values indicate stronger anomaly evidence, and the rubric defines the interpretation of each score range to ensure a consistent evaluation scale across datasets. 
(3) \textbf{Target node information:} this part provides node-specific content. For contextual anomaly detection, the LLM receives the target node's text attribute and the textual descriptions of its direct neighbors. For structural anomaly detection, the LLM receives a textualized subgraph representation that encodes relational or local edge information. Formally, the complete input to the LLM is:
\begin{equation}
\begin{aligned}
Score(x_i) &= \text{LLM}\big(
S \oplus UserPrompt \oplus \\
&\quad NodeInfo(x_i, \mathcal{N}_i(x_i))
\big)
\end{aligned}
\end{equation}

where $S$ denotes the analysis framework, \texttt{UserPrompt} encodes the scoring rubric and the task description, and \texttt{NodeInfo} contains textualized node and neighborhood information. The LLM produces both a natural language rationale and a final integer score in the format:
\begin{center}
\texttt{Analysis: ...\\ RESULT: <score>}
\end{center}

This structured design ensures that reasoning is grounded in retrieved knowledge, guided by explicit evaluation criteria, and informed by local graph semantics. Importantly, no node labels are provided; zero-shot anomaly scores depend solely on the retrieved framework, the task rubric, and the node’s textual or structural context. Moreover, RAG-assisted prompting reduces reliance on manually engineered prompts and applies a unified analysis framework across all nodes, datasets, and anomaly types, improving robustness and consistency while remaining fully zero-shot.

\section{Experiments}
\label{sec:exper}

\subsection{Experimental Setup}
\subsubsection{Datasets}
\noindent We evaluate on four widely used TAG datasets: \textbf{Cora} \cite{mccallum2000automating}, \textbf{Citeseer} \cite{giles1998citeseer}, \textbf{PubMed} \cite{sen2008collective}, and \textbf{WikiCS} \cite{mernyei2020wiki}. Each dataset consists of scientific publications represented as nodes, citation links as edges, and textual metadata (titles or abstracts) as node attributes. Detailed information about these datasets is in Appendix \ref{appen:dataset}.

\noindent To construct TAG-AD, we inject synthetic anomalies into the graphs using the methods described in Section~\ref{subsec:anomaly_gen}. In addition, we combine LLM-generated anomalies with structural anomalies to form a mixed anomaly type in our experiments. For each dataset, we sample a fixed proportion of nodes (5\%) as anomaly candidates. In the case of mixed anomalies, approximately (2.5\%) of the nodes are sampled for LLM-generated anomalies and another (2.5\%) for structural anomalies.
Details about the different types of anomaly generation processes are provided in Appendix \ref{appen:ano_gen_details} and Appendix \ref{appen:gen_prompt}.

\subsubsection{Evaluation Methods}
We benchmark two categories of methods:

\noindent \textbf{GNN-based Anomaly Detectors.} We benchmark representative unsupervised GAD methods, including DOMINANT \cite{ding2019deep}, AnomalyDAE \cite{fan2020anomalydae}, CoLA \cite{liu2021anomaly}, CONAD \cite{chen2020generative}, DONE \cite{bandyopadhyay2020outlier} and AdONE \cite{bandyopadhyay2020outlier}.
    
\noindent \textbf{Zero-shot LLMs.} We evaluate four modern LLMs (Gemma-3-27b-it, GPT-4o-mini, DeepSeek-V3, Qwen3-14B) in a zero-shot setting. Each model is provided with the textual or structural information of the target node and its neighbors, and is asked to determine whether the target node is anomalous. The prompt design is model-agnostic, ensuring fair and consistent evaluation across all LLMs.

\subsubsection{Evaluation Metrics}
We adopt two widely used metrics for evaluating anomaly detection performance: 
\textbf{Area Under the Receiver Operating Characteristic Curve (ROC-AUC)} and \textbf{Average Precision (AP)}. 
ROC-AUC measures the probability that an anomalous node is ranked higher than a normal node, 
while AP summarizes the precision--recall tradeoff across different thresholds. 
Higher ROC-AUC and AP values indicate better detection performance.

\begin{table*}[t]
    \centering
    \caption{Performance comparison of unsupervised GAD methods and zero-shot LLMs across four TAG datasets with LLM-generated contextual anomalies. Best results for each category of methods are highlighted in \textbf{bold}.}
    \vspace{-0.1in}
    \label{tab::C_anomaly1}
    \renewcommand{\arraystretch}{0.8}
    \setlength{\tabcolsep}{4pt}
    \resizebox{\textwidth}{!}{
    \begin{tabular}{llcccccccc}
        \toprule
        \multirow{2}{*}{\textbf{Method}} & \multirow{2}{*}{\textbf{Models}} 
        & \multicolumn{2}{c}{\textbf{Cora}} 
        & \multicolumn{2}{c}{\textbf{Citeseer}} 
        & \multicolumn{2}{c}{\textbf{Pubmed}} 
        & \multicolumn{2}{c}{\textbf{WikiCS}} \\
        \cmidrule(lr){3-4} \cmidrule(lr){5-6} \cmidrule(lr){7-8} \cmidrule(lr){9-10}
        & & ROC-AUC $\uparrow$ & AP $\uparrow$ 
          & ROC-AUC $\uparrow$ & AP $\uparrow$ 
          & ROC-AUC $\uparrow$ & AP $\uparrow$ 
          & ROC-AUC $\uparrow$ & AP $\uparrow$ \\
        \midrule
        \multirow{6}{*}{Unsupervised GNNs-GAD} 
        & DOMINANT   & 0.550 & 0.060 & 0.568 & 0.060 & 0.515 & 0.051 & 0.526 & 0.054 \\
        & AnomalyDAE & 0.489 & 0.046 & 0.596 & 0.061 & 0.477 & 0.044 & 0.442 & 0.043 \\
        & CoLA       & \textbf{0.688} & \textbf{0.114} & \textbf{0.700} & \textbf{0.101} & \textbf{0.647} & \textbf{0.114} & \textbf{0.647} & \textbf{0.131} \\
        & DONE       & 0.466 & 0.049 & 0.479 & 0.047 & 0.527 & 0.053 & 0.482 & 0.050 \\
        & AdONE      & 0.549 & 0.065 & 0.550 & 0.054 & 0.545 & 0.055 & 0.570 & 0.057 \\
        & CONAD      & 0.547 & 0.060 & 0.549 & 0.056 & 0.537 & 0.053 & 0.529 & 0.055 \\
        \midrule
        \multirow{3}{*}{Zero-shot LLMs (with RAG)} 
        & Gemma-3-27b-it & 0.821 & 0.244 & 0.851 & 0.290 & 0.716 & 0.213 & 0.824 & 0.354 \\
        & GPT-4o-mini & 0.908 & 0.331 & 0.895 & 0.425 & 0.852 & 0.214 & 0.840 & 0.322 \\
        & Qwen-3-14B  & 0.939 & 0.374 & 0.928 & \textbf{0.441} & \textbf{0.892} & 0.427 & NA \footnotemark[1] & NA \footnotemark[1] \\
        & Deepseek-V3 & \textbf{0.966} & \textbf{0.523} & \textbf{0.943} & 0.412 & 0.885 & \textbf{0.470} & \textbf{0.919} & \textbf{0.402} \\
        \bottomrule
    \end{tabular}
    }
\end{table*}

\begin{table*}[t]
    \centering
    \caption{Performance comparison of unsupervised GNN-based anomaly detectors and zero-shot LLMs across four TAG datasets with structural anomalies. Best results for each category of methods are highlighted in \textbf{bold}.}
    \vspace{-0.1in}
    \label{tab::S_anomaly}
    \renewcommand{\arraystretch}{0.8}
    \setlength{\tabcolsep}{4pt}
    \resizebox{\textwidth}{!}{
    \begin{tabular}{llcccccccc}
        \toprule
        \multirow{2}{*}{\textbf{Method}} & \multirow{2}{*}{\textbf{Models}} 
        & \multicolumn{2}{c}{\textbf{Cora}} 
        & \multicolumn{2}{c}{\textbf{Citeseer}} 
        & \multicolumn{2}{c}{\textbf{Pubmed}} 
        & \multicolumn{2}{c}{\textbf{WikiCS}} \\
        \cmidrule(lr){3-4} \cmidrule(lr){5-6} \cmidrule(lr){7-8} \cmidrule(lr){9-10}
        & & ROC-AUC $\uparrow$ & AP $\uparrow$ 
          & ROC-AUC $\uparrow$ & AP $\uparrow$ 
          & ROC-AUC $\uparrow$ & AP $\uparrow$ 
          & ROC-AUC $\uparrow$ & AP $\uparrow$ \\
        \midrule
        \multirow{6}{*}{Unsupervised GNNs-GAD} 
        & DOMINANT   & 0.969 & 0.428 & 0.984 & 0.570 & 0.903 & 0.194 & 0.667 & 0.070 \\
        & AnomalyDAE & 0.449 & 0.044 & 0.521 & 0.048 & 0.471 & 0.043 & 0.478 & 0.044 \\
        & CoLA       & 0.677 & 0.095 & 0.830 & 0.166 & 0.709 & 0.121 & 0.582 & \textbf{0.072} \\
        & DONE       & \textbf{0.985} & \textbf{0.758} & \textbf{0.990} & \textbf{0.772} & 0.897 & 0.185 & 0.652 & 0.067 \\
        & AdONE      & 0.982 & 0.745 & 0.981 & 0.714 & \textbf{0.906} & \textbf{0.200} & 0.650 & 0.066 \\
        & CONAD      & 0.969 & 0.430 & 0.983 & 0.552 & 0.903 & 0.194 & \textbf{0.672} & \textbf{0.072} \\
        \midrule
        \multirow{3}{*}{Zero-shot LLMs (with RAG)} 
        & Gemma-3-27b-it   & 0.713 & 0.107 & 0.749 & 0.128 & 0.509 & 0.052 & 0.489 & 0.049 \\
        & GPT-4o-mini      & 0.347 & 0.042 & 0.490 & 0.051 & 0.278 & 0.035 & 0.445 & 0.043 \\
        & Qwen-3-14B       & 0.563 & 0.069 & 0.759 & 0.106 & 0.425 & 0.044 & NA \footnotemark[1] & NA \footnotemark[1] \\
        & Deepseek-V3 & \textbf{0.842} & \textbf{0.237} & \textbf{0.849} & \textbf{0.181} & \textbf{0.593} & \textbf{0.076} & \textbf{0.592} & \textbf{0.061} \\
        \bottomrule
    \end{tabular}
    }
    \vspace{-0.1in}
\end{table*}

\subsubsection{Implementation Details}
\label{subsubsec:imple_details}
\noindent For GNN-based methods, we follow the default implementations in PyGOD \cite{JMLR:v25:23-0963} and retrain each model using the default hyperparameter configurations. 

\noindent For zero-shot LLMs, we design two distinct encoding strategies—one designed to capture \textit{contextual information} and the other to capture \textit{structural information}—to formulate the model inputs. Additionally, we design three prompt templates that enable the LLMs to detect contextual, structural, or mixed anomalies. The two encoding strategies are defined as follows:

\noindent \textbf{Contextual encoding.} The contextual encoding strategy provides the textual context surrounding the target node. For each target node, we concatenate its text attribute followed by the text attributes of its 1-hop neighbors. If a node's text exceeds $m_{\text{token}}$ tokens, we truncate it to the first $m_{\text{token}}$ tokens. If the target node has more than $k$ first-hop neighbors, we randomly sample $k$ neighbors. In our experiments we set $m_{\text{token}}=1000$ and $k=20$ for the contextual prompt template and $m_{\text{token}}=1000$ and $k=10$ for the mixed prompt template.

\noindent \textbf{Structural Encoding.} The structural encoding strategy converts the topological information of the subgraph centered on the target node into a natural language description interpretable by LLMs. For each target node, we extract its 2-hop subgraph and apply the incident encoding method recommended by \emph{Talk Like a Graph}~\cite{fatemi2023talk} to translate the subgraph structure into textual form. In our experiments, we limit the encoding to at most $m_{\text{node}}=100$ nodes, and up to $m_{\text{incident}}=100$ incident relations per node for the structural prompt template, and to at most $m_{\text{node}}=50$ nodes with up to $m_{\text{incident}}=50$ incident relations per node for the mixed prompt template.

\noindent \textbf{Prompt Templates.} We design three types of prompt templates: contextual, structural and mixed, each tailored for detecting contextual anomalies, structural anomalies, and mixed anomalies, respectively. Each prompt template comprises three components: the task description with a scoring rubric, the analysis framework, and the node-specific information. For the node-specific information, at the test time, the contextual prompt template is supplied with contextually encoded text, the structural prompt template with structurally encoded text, and the mixed prompt template with both. 

\noindent \textbf{Analysis framework.} For most of our experiments, the RAG-generated analysis framework is employed. Concretely, we construct the corpus from \cite{ding2019deep}, \cite{Guti_rrez_G_mez_2020}, \cite{xu2025textattributedgraphanomalydetection}, and \cite{NEURIPS2022_acc1ec4a}, which serves as the knowledge base for the RAG module to synthesize analysis frameworks tailored to different prompt templates. The RAG system is implemented on top of PaperQA2\cite{skarlinski2024languageagentsachievesuperhuman}.
To assess the effectiveness of RAG in the context of LLM-based GAD, we additionally conduct an ablation experiment on both LLM-generated anomalies and structural anomalies, where we replace the RAG-derived analysis framework with either (i) a \textbf{plain prompt} (a minimal placeholder) or (ii) an \textbf{manual prompt} (manually crafted by a human with relevant knowledge of GAD). 
All the prompts used for zero-shot anomaly detection are provided in Appendix~\ref{appen:detect_prompt} and Appendix~\ref{appen:rag_prompt}, respectively.

\begin{table*}[t]
    \centering
    \caption{Performance comparison of unsupervised GNN-based anomaly detectors and zero-shot LLMs across four TAG datasets with mixed anomalies. Best results for each category of methods are highlighted in \textbf{bold}.}
    \vspace{-0.1in}
    \label{tab::M_anomaly}
    \renewcommand{\arraystretch}{0.8}
    \setlength{\tabcolsep}{4pt}
    \resizebox{\textwidth}{!}{
    \begin{tabular}{llcccccccc}
        \toprule
        \multirow{2}{*}{\textbf{Method}} & \multirow{2}{*}{\textbf{Models}} 
        & \multicolumn{2}{c}{\textbf{Cora}} 
        & \multicolumn{2}{c}{\textbf{Citeseer}} 
        & \multicolumn{2}{c}{\textbf{Pubmed}} 
        & \multicolumn{2}{c}{\textbf{WikiCS}} \\
        \cmidrule(lr){3-4} \cmidrule(lr){5-6} \cmidrule(lr){7-8} \cmidrule(lr){9-10}
        & & ROC-AUC $\uparrow$ & AP $\uparrow$ 
          & ROC-AUC $\uparrow$ & AP $\uparrow$ 
          & ROC-AUC $\uparrow$ & AP $\uparrow$ 
          & ROC-AUC $\uparrow$ & AP $\uparrow$ \\
        \midrule
        \multirow{6}{*}{Unsupervised GNNs-GAD} 
        & DOMINANT   & 0.710 & 0.176 & \textbf{0.761} & 0.247 & 0.713 & 0.103 & 0.589 & 0.060 \\
        & AnomalyDAE & 0.482 & 0.046 & 0.566 & 0.053 & 0.465 & 0.042 & 0.441 & 0.041 \\
        & CoLA       & 0.669 & 0.098 & 0.756 & 0.133 & 0.703 & \textbf{0.145} & \textbf{0.616} & \textbf{0.095} \\
        & DONE       & 0.710 & 0.360 & 0.704 & 0.282 & 0.715 & 0.103 & 0.572 & 0.060 \\
        & AdONE      & \textbf{0.740} & \textbf{0.478} & 0.738 & \textbf{0.293} & \textbf{0.742} & 0.114 & 0.595 & 0.060 \\
        & CONAD      & 0.715 & 0.177 & 0.751 & 0.235 & 0.723 & 0.105 & 0.589 & 0.060 \\
        \midrule
        \multirow{3}{*}{Zero-shot LLMs (with RAG)} 
        & Gemma-3-27b-it   & 0.604 & 0.075 & 0.541 & 0.052 & 0.496 & 0.051 & 0.537 & 0.049 \\
        & GPT-4o-mini      & 0.598 & 0.074 & 0.491 & 0.046 & \textbf{0.596} & \textbf{0.068} & \textbf{0.594} & 0.057 \\
        & Qwen-3-14B       & 0.646 & 0.082 & \textbf{0.596} & 0.066 & 0.447 & 0.046 & NA \footnotemark[1] & NA \footnotemark[1] \\
        & Deepseek-V3 & \textbf{0.656} & \textbf{0.097} & 0.590 & \textbf{0.071} & 0.465 & 0.049 & 0.580 & \textbf{0.063} \\
        \bottomrule
    \end{tabular}
    }
\end{table*}

\begin{figure*}[t]
    \centering
    \includegraphics[width=1.0\linewidth]{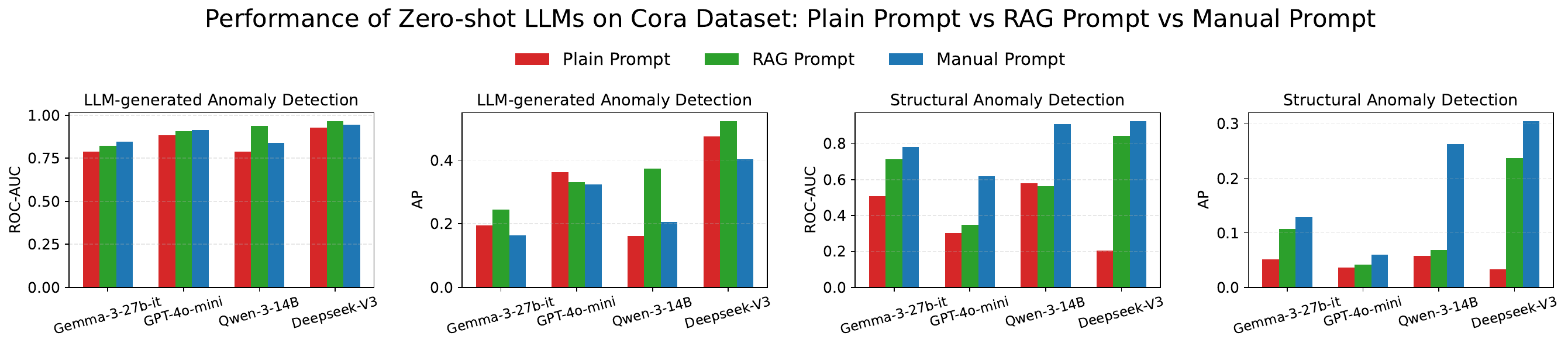}
    \caption{
        Performance comparison of zero-shot LLMs on the Cora dataset using Plain Prompt, RAG Prompt, and Manual Prompt. \textbf{Plain Prompt}: constructed without an explicit analysis framework. \textbf{RAG Prompt}: incorporates a retrieval-based analysis framework. \textbf{Manual Prompt}: uses a human-designed analysis framework.
    }
    \label{fig:prompts}
    \vspace{-0.1in}
\end{figure*}

\subsection{Main Results}
Results for detecting LLM-generated contextual anomalies and structural anomalies are shown in Tables \ref{tab::C_anomaly1} and \ref{tab::S_anomaly}, respectively. More results for the remaining anomaly types are in Appendix \ref{appen:results}. From the results, we make several key observations:

\footnotetext[1]{The context window of Qwen-3-14B is limited to 32k tokens, which makes it incompatible with the WikiCS dataset under the same experimental settings as the other models. All remaining models use a 128k-token context window.}

\noindent \textbf{LLMs excel at detecting contextual anomalies while GNNs are good at detecting structural anomalies.}
From the results, we can see that for contextual anomaly detection, in most cases, all the zero-shot LLM methods achieve significantly better detection performance compared with the best unsupervised GNN methods on the corresponding datasets. This shows that LLMs, with their strong semantic understanding and reasoning ability, can easily identify subtle inconsistencies between a node’s textual description and its surrounding context. In contrast, GNN-based models trained only on numerical embeddings tend to overfit to local graph smoothness and fail to recognize fine-grained semantic deviations in node attributes.
However, for structural anomalies, LLMs often fail to identify anomalies correctly even when explicitly provided with graph topology or verbalized structural cues. This suggests that current LLMs lack a robust understanding of relational structures such as neighborhoods, connectivity, or community-level consistency, which are essential for detecting topological irregularities. On the other hand, GNNs naturally leverage message passing and structural priors, making them much more effective for detecting abnormal edges or nodes that disrupt local graph structures.

\noindent \textbf{More advanced LLMs have better anomaly detection performance.}
Across all four datasets, LLMs with larger model sizes or more extensive training generally achieve higher ROC-AUC and AP scores than smaller LLMs. For example, DeepSeek-V3 performs the best or near-best on most datasets, while smaller models such as Gemma-3-27b-it yield comparatively lower performance. This trend suggests that model capacity and pretraining scale contribute to stronger zero-shot reasoning on textual anomalies. 

\noindent \textbf{Injecting anomalies in the feature space vs. the raw text space.}
Although GNN-based GAD methods can detect synthetic anomalies injected in the embedding space fairly well, as demonstrated in prior works \cite{ding2019deep, NEURIPS2022_acc1ec4a}, we find that many GNNs struggle to detect realistic LLM-generated contextual anomalies, even when they are semantically obvious. This suggests that applying pretrained language models only to obtain node embeddings and then training GNN-based detectors on top of these embeddings does not guarantee semantic awareness. The resulting models still fail to capture subtle textual inconsistencies, because the GNNs operate primarily on smoothed feature propagation rather than explicit reasoning about node semantics.
In contrast, zero-shot LLM-based approaches reason directly over natural language descriptions and can flexibly interpret diverse anomaly patterns without task-specific training. Nevertheless, their weakness in structural reasoning suggests an opportunity for hybrid methods—combining GNNs’ structural inductive biases with LLMs’ contextual understanding—to achieve a unified, generalizable graph anomaly detector.


        
        
        


\subsection{Mixed Anomaly Detection}
Table~\ref{tab::M_anomaly} reports the performance of unsupervised GNN-based anomaly detectors and zero-shot LLMs on the mixed-anomaly setting, where both contextual and structural anomalies are introduced. This task is more challenging than detecting a single anomaly type, since models must capture inconsistencies in both textual attributes and graph structure at the same time.

Among GNN-based approaches, \textbf{AdONE} achieves the strongest overall performance, obtaining the highest ROC-AUC on Cora and PubMed and the highest AP on Citeseer. \textbf{CoLA} also performs well, especially on WikiCS. These results show that methods that jointly consider structure and node attributes remain effective in settings where multiple anomaly sources coexist. In contrast, reconstruction-only methods such as \textbf{AnomalyDAE} perform poorly, indicating that mixed anomalies are not well detected by feature reconstruction alone.

Zero-shot LLMs show mixed results. Although they can detect some anomalies without any task-specific training, their overall performance is lower than that of GNN-based detectors. \textbf{DeepSeek-V3} achieves the best LLM performance on Cora and WikiCS but performs poorly on PubMed. In contrast, \textbf{GPT-4o-mini} performs the best among LLMs on PubMed, showing that LLMs can still detect a subset of mixed anomalies. However, across most datasets, zero-shot LLMs remain behind the strongest GNN baselines.

These results indicate that mixed anomalies continue to require strong structural modeling, and current zero-shot LLMs do not consistently capture both semantic and structural signals at the same time. GNN-based detectors therefore remain more reliable in this setting. A potential future direction is to combine graph neural encoders with LLM-based reasoning so that semantic understanding and topology-aware modeling can be integrated within a single framework.

\subsection{Effectiveness of RAG}
Fig. \ref{fig:prompts} compares the performance of zero-shot LLM-based anomaly detection between the plain prompt, the RAG prompt, and the manual prompt. 

For structural anomaly detection, both the manual prompt and the RAG prompt yield a substantial performance gain. This improvement arises because both prompts incorporate background knowledge drawn from the GAD literature, such as canonical definitions and diagnostic cues for structural irregularities, and explicitly integrate these insights into the detection prompt. Such prior knowledge enables the LLM to reason more effectively about what constitutes abnormal structural behavior, leading to more accurate judgments even in the absence of task-specific training. 
While the manual prompt delivers the strongest performance, the RAG prompt offers the clear practical advantage of avoiding manually engineered, expert-crafted instructions.
Empirically, DeepSeek-V3, the most capable model in our experimental setup, exhibits the most significant improvement, with both ROC-AUC and AP increasing sharply over the plain prompt when either enhanced prompt is used.
This finding suggests that more powerful LLMs are better able to exploit the background knowledge encoded in the prompts, and that the benefits of incorporating such prior knowledge become more pronounced as model capacity increases.

In contrast, for contextual anomalies, the performance gain of RAG over the plain prompt is less pronounced. This result is expected, as detecting contextual anomalies typically depends on commonsense and semantic reasoning, capabilities that modern LLMs already acquire through large-scale pretraining. As a result, the additional retrieved information yields only marginal benefits in this setting. Notably, human-designed prompts do not achieve the strongest overall performance, suggesting that manually crafted, domain-specific reasoning templates may be constrained by human bias and cognitive limitations. Overall, RAG-assisted prompting attains the best performance, albeit with a modest margin, indicating that our RAG design is both effective and practically advantageous. In particular, RAG prompting requires minimal manual prompt engineering, which enhances its practicality and adaptability in real-world scenarios.

In summary, RAG is particularly effective when detecting anomaly types that depend on explicit prior knowledge—such as structural or domain-specific irregularities—where background definitions play a crucial role. For more general contextual anomalies, where commonsense suffices, its contribution is smaller. These results suggest that retrieval-augmented prompting can greatly improve zero-shot GAD performance in knowledge-intensive settings while reducing dependence on hand-crafted manual prompts.

\section{Conclusion and Future Directions}
\label{sec:conclusion}
\vspace{-0.1in}
In this paper, we introduce \textbf{TAG-AD}, the first comprehensive benchmark for anomaly node detection on TAGs. Our framework leverages LLMs to generate realistic contextual anomalies directly in the raw text space, overcoming the limitations of feature-level perturbations that often degrade semantic coherence. TAG-AD also integrates text perturbation contextual and structural anomaly injection strategies, providing a unified and flexible platform for evaluating GAD methods under diverse conditions.

We benchmark both unsupervised GNN-based anomaly detectors and zero-shot LLMs. The results highlight that while conventional graph models capture structural irregularities effectively, modern LLMs—especially with RAG prompting—demonstrate strong zero-shot capabilities in detecting text-driven anomalies without any task-specific training. These findings reveal the emerging potential of foundation models as versatile anomaly detectors on graph-structured data.

In future work, we plan to extend TAG-AD to multi-modal graph settings (e.g., graphs with textual, visual, and temporal attributes) and to explore instruction-tuned graph–language models that can jointly reason over structure and semantics. We hope that the release of TAG-AD will catalyze further research on integrating graph learning and large language models for robust, interpretable, and label-efficient anomaly detection.


\bibliography{custom}

\newpage
\appendix
\section*{Appendix of TAG-AD}

\section{Datasets}
\label{appen:dataset}

\paragraph{Cora}
The Cora dataset \cite{mccallum2000automating} contains 2,708 scientific publications categorized into seven research topics: Case-Based Reasoning, Genetic Algorithms, Neural Networks, Probabilistic Methods, Reinforcement Learning, Rule Learning, and Theory. Each node represents a paper, and edges denote citation links between papers, forming a graph with 5,429 edges.

\paragraph{CiteSeer}
The CiteSeer dataset \cite{giles1998citeseer} consists of 3,186 scientific publications grouped into six research domains: Agents, Machine Learning, Information Retrieval, Databases, Human–Computer Interaction, and Artificial Intelligence. Each node corresponds to a paper, with textual features extracted from its title and abstract. The graph is constructed based on citation relationships among the publications.

\paragraph{PubMed}
The PubMed dataset \cite{sen2008collective} comprises scientific articles related to diabetes research, divided into three categories: experimental studies on mechanisms and treatments, research on Type 1 Diabetes focusing on autoimmune factors, and Type 2 Diabetes studies emphasizing insulin resistance and management. Nodes represent papers, edges denote citation links, and node features are derived from the medical abstracts.

\paragraph{WikiCS}
The WikiCS dataset \cite{mernyei2020wiki} is a Wikipedia-based citation graph built for benchmarking graph neural networks. Nodes correspond to computer science articles categorized into ten subfields serving as class labels, while edges represent hyperlinks between articles. Node features are extracted from the textual content of the corresponding Wikipedia entries.

\section{GNN-based Unsupervised GAD Methods}
\label{appen:method}

\paragraph{DOMINANT \cite{ding2019deep}} DOMINANT is among the first approaches to combine GCNs with autoencoders for GAD. It employs a two-layer GCN encoder and two decoders: one reconstructs node attributes using a GCN, and the other reconstructs the adjacency matrix via a dot-product layer. The final anomaly score is the combination of reconstruction errors from both decoders.

\paragraph{DONE \cite{bandyopadhyay2020outlier}} DONE uses separate autoencoders for structural and attribute reconstruction, both implemented with multilayer perceptrons. The method jointly optimizes node embeddings and anomaly scores through a unified loss that balances reconstruction consistency.

\paragraph{AdONE \cite{bandyopadhyay2020outlier}} AdONE extends DONE by introducing an adversarial discriminator to align structure and attribute embeddings in the latent space. The discriminator encourages consistency between the two representations, leading to more robust detection performance.

\paragraph{CoLA \cite{liu2021anomaly}} CoLA employs a contrastive learning framework that captures informative representations by contrasting node–subgraph pairs. The learned representations enable the model to assign discriminative anomaly scores, allowing effective ranking of abnormal nodes.

\paragraph{AnomalyDAE \cite{fan2020anomalydae}} AnomalyDAE leverages dual autoencoders for structure and attributes. Its structural encoder jointly processes the adjacency matrix and node features, while the attribute decoder reconstructs the node attributes using both structural and attribute embeddings. The reconstruction error serves as the node’s anomaly score.

\paragraph{CONAD \cite{chen2020generative}} CONAD belongs to the family of BOND-based methods that integrate graph augmentation with contrastive learning. It introduces prior knowledge about potential outlier nodes by generating augmented graph views. These augmented graphs are encoded using Siamese GNN encoders, and the model is optimized via a contrastive loss to enhance representation consistency. Similar to DOMINANT, the final anomaly score for each node is obtained through two separate decoders that reconstruct structural and attribute information.

\label{appen:results}
\begin{table*}[t]
    \centering
    \caption{Performance comparison of unsupervised GNN-based anomaly detectors and zero-shot LLMs across four TAG datasets with global contextual anomalies. Best results are highlighted in \textbf{bold}.}
    \label{tab::G_anomaly}
    \renewcommand{\arraystretch}{1.0}
    \setlength{\tabcolsep}{4pt}
    \resizebox{\textwidth}{!}{
    \begin{tabular}{llcccccccc}
        \toprule
        \multirow{2}{*}{\textbf{Method}} & \multirow{2}{*}{\textbf{Models}} 
        & \multicolumn{2}{c}{\textbf{Cora}} 
        & \multicolumn{2}{c}{\textbf{Citeseer}} 
        & \multicolumn{2}{c}{\textbf{Pubmed}} 
        & \multicolumn{2}{c}{\textbf{WikiCS}} \\
        \cmidrule(lr){3-4} \cmidrule(lr){5-6} \cmidrule(lr){7-8} \cmidrule(lr){9-10}
        & & ROC-AUC $\uparrow$ & AP $\uparrow$ 
          & ROC-AUC $\uparrow$ & AP $\uparrow$ 
          & ROC-AUC $\uparrow$ & AP $\uparrow$ 
          & ROC-AUC $\uparrow$ & AP $\uparrow$ \\
        \midrule
        \multirow{6}{*}{Unsupervised GNNs-GAD} 
        & DOMINANT   & 0.597 & 0.070 & 0.602 & 0.064 & 0.654 & 0.069 & 0.528 & 0.054 \\
        & AnomalyDAE & \textbf{0.952} & \textbf{0.431} & \textbf{0.734} & \textbf{0.087} & 0.299 & 0.033 & 0.409 & 0.039 \\
        & CoLA       & 0.589 & 0.075 & 0.665 & 0.086 & 0.651 & \textbf{0.119} & 0.633 & \textbf{0.111} \\
        & DONE       & 0.553 & 0.063 & 0.549 & 0.052 & \textbf{0.778} & 0.097 & \textbf{0.651} & 0.067 \\
        & AdONE      & 0.511 & 0.053 & 0.524 & 0.049 & 0.758 & 0.090 & 0.614 & 0.061 \\
        & CONAD      & 0.603 & 0.071 & 0.593 & 0.062 & 0.651 & 0.068 & 0.531 & 0.056 \\
        \midrule
        \multirow{3}{*}{Zero-shot LLMs (with RAG)} 
        & Gemma-3-27b-it    & 0.984 & 0.678 & 0.927 & 0.553 & 0.977 & 0.730 & 0.895 & 0.400 \\
        & GPT-4o-mini       & 0.983 & 0.696 & 0.915 & 0.547 & 0.976 & 0.627 & 0.880 & 0.298 \\
        & Qwen-3-14B        & 0.985 & 0.637 & 0.946 & 0.595 & 0.983 & 0.675 & NA \footnotemark[1] & NA \footnotemark[1] \\
        & Deepseek-V3 & \textbf{0.993} & \textbf{0.800} & \textbf{0.963} & \textbf{0.624} & \textbf{0.992} & \textbf{0.781} & \textbf{0.953} & \textbf{0.544} \\
        \bottomrule
    \end{tabular}
    }
\end{table*}

\begin{table*}[t]
    \centering
    \caption{Performance comparison of unsupervised GNN-based anomaly detectors and zero-shot LLMs across four TAG datasets with text perturbation contextual anomalies. Best results are highlighted in \textbf{bold}.}
    \label{tab::TC_anomaly}
    \renewcommand{\arraystretch}{1.0}
    \setlength{\tabcolsep}{4pt}
    \resizebox{\textwidth}{!}{
    \begin{tabular}{llcccccccc}
        \toprule
        \multirow{2}{*}{\textbf{Method}} & \multirow{2}{*}{\textbf{Models}} 
        & \multicolumn{2}{c}{\textbf{Cora}} 
        & \multicolumn{2}{c}{\textbf{Citeseer}} 
        & \multicolumn{2}{c}{\textbf{Pubmed}} 
        & \multicolumn{2}{c}{\textbf{WikiCS}} \\
        \cmidrule(lr){3-4} \cmidrule(lr){5-6} \cmidrule(lr){7-8} \cmidrule(lr){9-10}
        & & ROC-AUC $\uparrow$ & AP $\uparrow$ 
          & ROC-AUC $\uparrow$ & AP $\uparrow$ 
          & ROC-AUC $\uparrow$ & AP $\uparrow$ 
          & ROC-AUC $\uparrow$ & AP $\uparrow$ \\
        \midrule
        \multirow{6}{*}{Unsupervised GNNs-GAD} 
        & DOMINANT   & 0.500 & 0.052 & 0.498 & 0.049 & 0.510 & 0.051 & 0.507 & 0.051 \\
        & AnomalyDAE & 0.445 & 0.043 & 0.547 & 0.062 & 0.483 & 0.047 & 0.530 & 0.055 \\
        & CoLA      & \textbf{0.564} & \textbf{0.066} & \textbf{0.598} & \textbf{0.068} & \textbf{0.633} & \textbf{0.105} & \textbf{0.560} & \textbf{0.081} \\
        & DONE      & 0.481 & 0.056 & 0.486 & 0.046 & 0.555 & 0.055 & 0.503 & 0.051 \\
        & AdONE     & 0.484 & 0.051 & 0.488 & 0.047 & 0.540 & 0.054 & 0.492 & 0.049 \\
        & CONAD     & 0.500 & 0.053 & 0.493 & 0.048 & 0.519 & 0.052 & 0.509 & 0.052 \\
        \midrule
        \multirow{3}{*}{Zero-shot LLMs (with RAG)} 
        & Gemma-3-27b-it    & 0.742 & 0.168 & 0.769 & \textbf{0.186} & 0.816 & \textbf{0.391} & 0.638 & \textbf{0.081} \\
        & GPT-4o-mini       & 0.619 & 0.082 & 0.666 & 0.074 & 0.602 & 0.057 & 0.586 & 0.055 \\
        & Qwen-3-14B        & 0.758 & 0.162 & \textbf{0.812} & 0.146 & 0.835 & 0.293 & NA \footnotemark[1] & NA \footnotemark[1] \\
        & Deepseek-V3  & \textbf{0.797} & \textbf{0.234} & 0.781 & 0.161 & \textbf{0.841} & 0.328 & \textbf{0.654} & 0.078 \\
        \bottomrule
    \end{tabular}
    }
\end{table*}

\section{More Results}
\label{appen:results}
We further present anomaly detection results for both global contextual and text perturbation anomalies. As shown in Tables \ref{tab::G_anomaly} and \ref{tab::TC_anomaly}, in the case of contextual anomaly detection, the zero-shot LLM-based methods generally outperform the best unsupervised GNN baselines across most datasets. This demonstrates that LLMs, leveraging their strong semantic understanding and reasoning capabilities, can effectively capture subtle inconsistencies between a node’s textual description and its surrounding context.

\section{Anomaly Generation Details}
\label{appen:ano_gen_details}
\noindent The detailed procedure for constructing the LLM-generated contextual anomalies is described in Section~\ref{subsubsec:imple_details}. For text perturbation anomalies, we set $K_{\text{candidate}} = 50$ in our experiments. The replaced sequence length $l$ is computed as:
\[
l = \min\bigl(l_s \times a,\; l_t\bigr), \quad a \sim \mathcal{U}(0.25,\, 0.5),
\]
where $l_t$ and $l_s$ denote the numbers of tokens in the target and source node text attributes, respectively. Regarding global contextual anomalies, we employ text attributes from \textbf{WikiCS} as the outlier source for \textbf{Cora} and \textbf{Citeseer}, and use \textbf{Citeseer} as the outlier source for \textbf{WikiCS}. The outlier text source for \textbf{PubMed} is built by combining four PubMed datasets \footnotemark[2] from HuggingFace. For structural anomalies, we utilize PyGOD~\cite{JMLR:v25:23-0963} with the parameter $m = 10$.

\footnotetext[2]{Datasets used for constructing global contextual anomalies for Pubmed Datasets: \\
\url{https://huggingface.co/datasets/Gaborandi/breast_cancer_pubmed_abstracts} \\
\url{https://huggingface.co/datasets/Gaborandi/HIV_pubmed_abstracts} \\
\url{https://huggingface.co/datasets/Gaborandi/Brain_Tumor_pubmed_abstracts} \\
\url{https://huggingface.co/datasets/Gaborandi/Alzheimer_pubmed_abstracts}
}

\section{Anomaly Generation Prompts}
\label{appen:gen_prompt}

\begin{promptbox}
\begin{Verbatim}[
    breaklines,                
    breakanywhere,
    showtabs=false, showspaces=false,
    breaksymbol={},
    formatcom=\ttfamily\small
]
SYSTEM_PROMPT = """
You are a TAG (Text-Attributed Graph) anomaly text generator. Your task is to generate contextual anomalies by rewriting node text attributes. Strictly adhere to all provided constraints and requirements.
"""
\end{Verbatim}
\end{promptbox}

\begin{promptbox}
\begin{Verbatim}[
    breaklines,                
    breakanywhere,
    showtabs=false, showspaces=false,
    breaksymbol={},
    formatcom=\ttfamily\small
]
USER_PROMPT_CORA = """
You are given a normal node from the Cora dataset. The Cora dataset contains Machine Learning publications categorized into seven topics:
1. Machine Learning: Case-Based
2. Machine Learning: Genetic Algorithms
3. Machine Learning: Neural Networks
4. Machine Learning: Probabilistic Methods
5. Machine Learning: Reinforcement Learning
6. Machine Learning: Rule Learning
7. Machine Learning: Theory
Each node's text attribute is the paper title and, optionally, an abstract.

Original node label/topic: ({label_name})
Designated target label/topic: ({designated_label})
Original text attribute: {raw_text}

Task: Rewrite the text attribute so it clearly concerns the designated label/topic: ({designated_label}) while remaining a **contextual anomaly** relative to the original.

Requirements:
1) **Topicality**: The rewritten text must align tightly with ({designated_label}).
2) **Divergence**: Make the content unrelated to the original topic ({label_name}) as much as possible.
3) **Style match**: Mimic the original tone, formality, and sentence structure; keep the length of the rewritten text roughly similar to the original.
4) **Structure preservation**: If the original text only includes an abstract, output only an abstract; if it includes a title and an abstract, output a title and an abstract.
5) **No extras**: Do not add author names, citations, URLs, dataset commentary, or title/abstract labels (e.g., "Title: ", "Abstract: ").
6) **Output format**: Return only the rewritten text, formatted exactly like the original text attribute. Do not include any provenance/generation markers (e.g., "Rewritten text: "). The output should read naturally and contain no phrasing that suggests it was produced by a language model.
"""
\end{Verbatim}
\end{promptbox}

\begin{promptbox}
\begin{Verbatim}[
    breaklines,                
    breakanywhere,
    showtabs=false, showspaces=false,
    breaksymbol={},
    formatcom=\ttfamily\small
]
USER_PROMPT_CITESEER = """
You are given a normal node from the CiteSeer dataset.
The CiteSeer dataset contains Computer Science publications categorized into six topics:
1. Computer Science: Agents
2. Computer Science: Machine Learning
3. Computer Science: Information Retrieval
4. Computer Science: Database
5. Computer Science: Human Computer Interaction
6. Computer Science: Artificial Intelligence
Each node's text attribute is the paper abstract.

Original node label/topic: ({label_name})
Designated target label/topic: ({designated_label})
Original text attribute: {raw_text}

Task: Rewrite the text attribute so it clearly concerns the designated label/topic: ({designated_label}) while remaining a **contextual anomaly** relative to the original.

Requirements:
1) **Topicality**: The rewritten text must align tightly with ({designated_label}).
2) **Divergence**: Make the content unrelated to the original topic ({label_name}) as much as possible.
3) **Style match**: Mimic the original tone, formality, and sentence structure; keep the length of the rewritten text roughly similar to the original.
4) **Structure preservation**: If the original text only includes an abstract, output only an abstract; if it includes a title and an abstract, output a title and an abstract.
5) **No extras**: Do not add author names, citations, URLs, dataset commentary, or title/abstract labels (e.g., "Title: ", "Abstract: ").
6) **Output format**: Return only the rewritten text, formatted exactly like the original text attribute. Do not include any provenance/generation markers (e.g., "Rewritten text: ").
The output should read naturally and contain no phrasing that suggests it was produced by a language model.
"""
\end{Verbatim}
\end{promptbox}

\begin{promptbox}
\begin{Verbatim}[
    breaklines,                
    breakanywhere,
    showtabs=false, showspaces=false,
    breaksymbol={},
    formatcom=\ttfamily\small
]
USER_PROMPT_PUBMED = """
You are given a normal node from the PubMed dataset.
The PubMed dataset contains Diabetes Mellitus publications categorized into three topics:
1. Diabetes Mellitus, Experimental
2. Diabetes Mellitus Type 1
3. Diabetes Mellitus Type 2
Each node's text attribute is the paper title and abstract.

Original node label/topic: ({label_name})
Designated target label/topic: ({designated_label})
Original text attribute: {raw_text}

Task: Rewrite the text attribute so it clearly concerns the designated label/topic: ({designated_label}) while remaining a **contextual anomaly** relative to the original.

Requirements:
1) **Topicality**: The rewritten text must align tightly with ({designated_label}).
2) **Divergence**: Make the content unrelated to the original topic ({label_name}) as much as possible.
3) **Style match**: Mimic the original tone, formality, and sentence structure; keep the length of the rewritten text roughly similar to the original.
4) **Structure preservation**: If the original text only includes an abstract, output only an abstract; if it includes a title and an abstract, output a title and an abstract.
5) **No extras**: Do not add author names, citations, URLs, or dataset commentary.
6) **Output format**: Return only the rewritten text, formatted exactly like the original text attribute. Do not include any provenance/generation markers (e.g., "Rewritten text: "). The output should read naturally and contain no phrasing that suggests it was produced by a language model.
"""
\end{Verbatim}
\end{promptbox}

\begin{promptbox}
\begin{Verbatim}[
    breaklines,                
    breakanywhere,
    showtabs=false, showspaces=false,
    breaksymbol={},
    formatcom=\ttfamily\small
]
USER_PROMPT_WIKICS = """
You are given a normal node from the Wiki-CS dataset.
The Wiki-CS dataset contains Computer Science articles from Wikipedia categorized into ten topics:
1. Computational linguistics
2. Databases
3. Operating systems
4. Computer architecture
5. Computer security
6. Internet protocols
7. Computer file systems
8. Distributed computing architecture
9. Web technology
10. Programming language topics
Each node's text attribute is the text of the article.

Original node label/topic: ({label_name})
Designated target label/topic: ({designated_label})
Original text attribute: {raw_text}

Task: Rewrite the text attribute so it clearly concerns the designated label/topic: ({designated_label}) while remaining a **contextual anomaly** relative to the original.

Requirements:
1) **Topicality**: The rewritten text must align tightly with ({designated_label}).
2) **Divergence**: Make the content unrelated to the original topic ({label_name}) as much as possible.
3) **Style match**: Mimic the original tone, formality, and sentence structure; keep the length of the rewritten text roughly similar to the original.
4) **Structure preservation**: If the original text only includes a title, output only a title; if it includes a title and a text, output a title and a text.
5) **No extras**: Do not add citations, URLs, dataset commentary, or title/text labels (e.g., "Title: ", "Text: ").
6) **Output format**: Return only the rewritten text, formatted exactly like the original text attribute. Do not include any provenance/generation markers (e.g., "Rewritten text: "). The output should read naturally and contain no phrasing that suggests it was produced by a language model.
"""
\end{Verbatim}
\end{promptbox}

\section{Zero-shot Detection Prompts}
\label{appen:detect_prompt}

\begin{promptbox}
\begin{Verbatim}[
    breaklines,                
    breakanywhere,
    showtabs=false, showspaces=false,
    breaksymbol={},
    formatcom=\ttfamily\small
]
SYSTEM_PROMPT = """
You are an expert anomaly detector specializing in Text-Attributed Graphs (TAGs)
"""
\end{Verbatim}
\end{promptbox}

\begin{promptbox}
\begin{Verbatim}[
    breaklines,                
    breakanywhere,
    showtabs=false, showspaces=false,
    breaksymbol={},
    formatcom=\ttfamily\small
]
NEIGHBORS_DESCRIPTION = """
**Neighbor {neighbor_index}**
Text attribute: {neighbor_text}

"""
\end{Verbatim}
\end{promptbox}

\begin{promptbox}
\begin{Verbatim}[
    breaklines,                
    breakanywhere,
    showtabs=false, showspaces=false,
    breaksymbol={},
    formatcom=\ttfamily\small
]
USER_PROMPT_CONTEXTUAL = """
You are analyzing a node from a text-attributed graph for anomaly detection. You will be given the text attribute of the target node \
and the text attributes of its direct neighbors. The text attributes are truncated to {max_word_num} words to reduce the length of the prompt.

**Target Node Text**
{raw_text}

**Description of {num_neighbors} direct neighbors**

{neighbors_description}
Task:
Predict an integer anomaly score (0-10) for the target node by following the analysis framework.

Analysis Framework:
{analysis_framework}

Rubric:
The score should be an integer from 0 to 10 (higher = stronger anomaly evidence)
**Score 0**: definitely normal
**Score 1-4**: more likely normal (minor issues)
**Score 5**: equally likely normal and anomalous
**Score 6-9**: more likely anomalous (increasing evidence)
**Score 10**: definitely anomalous

Output Format:
1) Start with a detailed analysis following the analysis framework
2) Conclude with "RESULT:<score>". Score should be an integer from 0 to 10. For example, "RESULT:0"
3) The result score must appear on its own line at the very end of your response

Output Example:
Analysis: This given node has ...
RESULT:5
"""
\end{Verbatim}
\end{promptbox}

\begin{promptbox}
\begin{Verbatim}[
    breaklines,                
    breakanywhere,
    showtabs=false, showspaces=false,
    breaksymbol={},
    formatcom=\ttfamily\small
]
USER_PROMPT_STRUCTURAL = """
You are analyzing a node from a text-attributed graph for structural anomaly detection. You will receive a textual \
representation describing the structure of a subgraph centered around the target node {idx}.

**Subgraph Structure Representation**
{graph_structure_representation}

Task:
Predict an integer anomaly score (0-10) for the target node by following the analysis framework.

Analysis Framework:
{analysis_framework}

Rubric:
The score should be an integer from 0 to 10 (higher = stronger anomaly evidence)
**Score 0**: definitely normal
**Score 1-4**: more likely normal (minor issues)
**Score 5**: equally likely normal and anomalous
**Score 6-9**: more likely anomalous (increasing evidence)
**Score 10**: definitely anomalous

Output Format:
1) Start with a detailed analysis following the analysis framework
2) Conclude with "RESULT:<score>". Score should be an integer from 0 to 10. For example, "RESULT:0"
3) The result score must appear on its own line at the very end of your response

Output Example:
Analysis: This given node has ...
RESULT:5
"""
\end{Verbatim}
\end{promptbox}

\begin{promptbox}
\begin{Verbatim}[
    breaklines,                
    breakanywhere,
    showtabs=false, showspaces=false,
    breaksymbol={},
    formatcom=\ttfamily\small
]
USER_PROMPT_MIXED = """
You are analyzing a node from a text-attributed graph for anomaly detection. You will be given:
1) The text attribute of the target node, and the text attributes of its direct neighbors. The text attributes are truncated to {max_word_num} words to reduce the length of the prompt.
2) A textual representation describing the structure of a 2-hop subgraph centered at the target node {idx}.

**Target Node Text**
{raw_text}

**Description of {num_neighbors} direct neighbors**

{neighbors_description}

**Subgraph Structure Representation**
{graph_structure_representation}

Task:
Predict an integer anomaly score (0-10) for the target node by following the analysis framework.

Analysis Framework:
{analysis_framework}

Rubric:
The score should be an integer from 0 to 10 (higher = stronger anomaly evidence)
**Score 0**: definitely normal
**Score 1-4**: more likely normal (minor issues)
**Score 5**: equally likely normal and anomalous
**Score 6-9**: more likely anomalous (increasing evidence)
**Score 10**: definitely anomalous

Output Format:
1) Start with a detailed analysis following the analysis framework
2) Conclude with "RESULT:<score>". Score should be an integer from 0 to 10. For example, "RESULT:0"
3) The result score must appear on its own line at the very end of your response

Output Example:
Analysis: This given node has ...
RESULT:5
"""
\end{Verbatim}
\end{promptbox}

\begin{promptbox}
\begin{Verbatim}[
    breaklines,                
    breakanywhere,
    showtabs=false, showspaces=false,
    breaksymbol={},
    formatcom=\ttfamily\small
]
ANALYSIS_FRAMEWORK_CONTEXTUAL_HUMAN_DESIGNED = """An contextual anomaly node exhibits one or more of the following characteristics:
1) **Content corruption**: The text attribute contains corrupted, nonsensical, low-coherence, spammy, or irrelevant text.
2) **Neighbor inconsistency**: Weak semantic relatedness with the majority of direct neighbors; off-topic vs. local neighborhood themes.
3) **Contextual inappropriateness**: The node's content is contextually inappropriate for its position in the graph structure.
You should analyze each of the above characteristics from the following aspects:
1) **Quality Assessment**: Evaluate the quality and coherence of the target node's text attribute.
2) **Neighbor Coherence**: Assess semantic similarity and topical consistency with direct neighbors.
3) **Graph Context**: Judge whether the node fits naturally within its local graph neighborhood."""
\end{Verbatim}
\end{promptbox}

\begin{promptbox}
\begin{Verbatim}[
    breaklines,                
    breakanywhere,
    showtabs=false, showspaces=false,
    breaksymbol={},
    formatcom=\ttfamily\small
]
ANALYSIS_FRAMEWORK_STRUCTURAL_HUMAN_DESIGNED = """A structural anomaly node exhibits one or more of the following characteristics:
1) **Clique-like density spike**: The node sits inside an unusually dense (near-)clique; its immediate neighbors are also heavily interconnected.
2) **Egonet surplus vs. expectation**: The node's egonet (the node and its immediate neighbors, plus all edges among them) contains far more edges or triangles than would be expected for a node of its degree or position in the graph.
3) **Boundary sparsity**: The dense core around the node has relatively few edges that cross to the outside, creating a sharp contrast between internal density and external connectivity.
You should analyze each of the above characteristics from the following aspects:
1) **Local Structural Intensity**: Measure the internal connectivity around the target node (e.g., number of edges, triangles, or density within the egonet). Compare these values to local baselines, such as the egonets of neighbors with similar degree or position.
2) **Boundary & Cut Properties**: Assess the number of edges that connect the egonet (the node and its immediate neighbors) to the rest of the graph. Determine whether there is a sharp drop in connectivity at the boundary, indicating a well-separated dense core.
3) **Community & Positional Consistency**: Evaluate whether the node's structural position and its local community structure are consistent with the broader graph. Consider if the node is embedded in a community in a way that is unusual or inconsistent with typical nodes in the graph."""
\end{Verbatim}
\end{promptbox}

\begin{promptbox}
\begin{Verbatim}[
    breaklines,
    breakanywhere,
    showtabs=false, showspaces=false,
    breaksymbol={},
  formatcom=\ttfamily\small
]
ANALYSIS_FRAMEWORK_PLACEHOLDER = """
You should design an analysis framework by yourself. Then follow your analysis framework to analyze the target node.
"""
\end{Verbatim}
\end{promptbox}

\section{RAG Prompts}
\label{appen:rag_prompt}

\begin{promptbox}
\begin{Verbatim}[
    breaklines,                
    breakanywhere,
    showtabs=false, showspaces=false,
    breaksymbol={},
    formatcom=\ttfamily\small
]
ANALYSIS_FRAMEWORK_EXAMPLE = """An <type_of_anomaly> anomaly node exhibits one or more of the following characteristics:
1) **<characteristic_1>**: 1 sentences
2) **<characteristic_2>**: 1 sentences
...
You should analyze each of the above characteristics from the following aspects:
1) **<aspect_1>**: 1 sentences
2) **<aspect_2>**: 1 sentences
..."""
\end{Verbatim}
\end{promptbox}

\begin{promptbox}
\begin{Verbatim}[
    breaklines,                
    breakanywhere,
    showtabs=false, showspaces=false,
    breaksymbol={},
    formatcom=\ttfamily\small
]
CONTEXTUAL_ANOMALY_PROMPT_RAG = """
You task is to generate the Analysis Framework Section of a prompt used to detect contextual anomaly.
The Analysis Framework should be concise, practical, and directly usable within an LLM prompt for anomaly detection. Do not include any instructions that go beyond the scope of the inputs. Do not include any instructions that cannot be executed by the large language model(e.g., compare embeddings).
The ONLY inputs available to the model at inference time: the target node's text and the texts of its direct neighbors.
The prompt has already contained the following Rubric Section:

Rubric:
The score should be an integer from 0 to 10 (higher = stronger anomaly evidence)
**Score 0**: definitely normal
**Score 1-4**: more likely normal (minor issues)
**Score 5**: equally likely normal and anomalous
**Score 6-9**: more likely anomalous (increasing evidence)
**Score 10**: definitely anomalous

You should generate the Analysis Framework Section for the contextual anomaly without any other information such as titles, citations, references, summaries, conclusions, etc.
Here is an example:
{example}
"""
\end{Verbatim}
\end{promptbox}

\begin{promptbox}
\begin{Verbatim}[
    breaklines,                
    breakanywhere,
    showtabs=false, showspaces=false,
    breaksymbol={},
    formatcom=\ttfamily\small
]
STRUCTURAL_ANOMALY_PROMPT_RAG = """
You task is to generate the Analysis Framework Section of a prompt used to detect structural anomaly.
The Analysis Framework should be concise, practical, and directly usable within an LLM prompt for anomaly detection.
Do not include any instructions that go beyond the scope of the inputs. Do not include any instructions that cannot be executed by the large language model(e.g., compare embeddings).
The ONLY inputs available to the model at inference time: the structure of a 2-hop subgraph centered at the target node. The text attribute of the target node is not available.
The prompt has already contained the following Rubric Section:

Rubric:
The score should be an integer from 0 to 10 (higher = stronger anomaly evidence)
**Score 0**: definitely normal
**Score 1-4**: more likely normal (minor issues)
**Score 5**: equally likely normal and anomalous
**Score 6-9**: more likely anomalous (increasing evidence)
**Score 10**: definitely anomalous

You should generate the Analysis Framework Section for the contextual anomaly without any other information such as titles, citations, references, summaries, conclusions, etc.
Here is an example:
{example}
"""
\end{Verbatim}
\end{promptbox}

\begin{promptbox}
\begin{Verbatim}[
    breaklines,
    breakanywhere,
    showtabs=false, showspaces=false,
    breaksymbol={},
    formatcom=\ttfamily\small
]
MIXED_ANOMALY_PROMPT_RAG = """
You task is to generate the Analysis Framework Section of a prompt used to detect both contextual and structural anomaly.
The Analysis Framework should be concise, practical, and directly usable within an LLM prompt for anomaly detection.
Do not include any instructions that go beyond the scope of the inputs. Do not include any instructions that cannot be executed by the large language model(e.g., compare embeddings).
The ONLY inputs available to the model at inference time:
1) The text attributes of the target node and its direct neighbors.
2) The structure of a 2-hop subgraph centered at the target node.
The prompt has already contained the following Rubric Section:

Rubric:
The score should be an integer from 0 to 10 (higher = stronger anomaly evidence)
**Score 0**: definitely normal
**Score 1-4**: more likely normal (minor issues)
**Score 5**: equally likely normal and anomalous
**Score 6-9**: more likely anomalous (increasing evidence)
**Score 10**: definitely anomalous

You should generate the Analysis Framework Section for the mixed anomaly without any other information such as titles, citations, references, summaries, conclusions, etc.
Here is an example:
{example}
"""
\end{Verbatim}
\end{promptbox}
\end{document}